\documentclass{article}
\usepackage{subcaption}

\usepackage{times}  % DO NOT CHANGE THIS
\usepackage{helvet} % DO NOT CHANGE THIS
\usepackage{courier}  % DO NOT CHANGE THIS
\usepackage[hyphens]{url}  % DO NOT CHANGE THIS
\usepackage{graphicx} % DO NOT CHANGE THIS
\usepackage{caption} % DO NOT CHANGE THIS AND DO NOT ADD ANY OPTIONS TO IT
\usepackage{authblk}

\usepackage[ruled,vlined]{algorithm2e}
\usepackage[utf8]{inputenc}
\usepackage{amsfonts}
\usepackage{amsmath}
\usepackage[T1]{fontenc}

\title{Using Under-trained Deep Ensembles to Learn Under Extreme Label Noise}

\author[1]{\small Konstantinos Nikolaidis \thanks{Corresponding Author: konstan@ifi.uio.no}}

\author[1]{\small Thomas Plagemann}
\author[1]{\small Stein Kristiansen}

\author[1]{\small Vera Goebel}
\author[2]{\small Mohan Kankanhalli}

\affil[1]{\scriptsize Department of Informatics, University of Oslo, Norway}

\affil[2]{Department of Computer Science, National University of Singapore, Singapore}

\begin{document}
\frenchspacing

\maketitle
\begin{abstract}
    
Improper or erroneous labelling can pose a hindrance to reliable generalization for supervised learning. This can have negative consequences, especially for critical fields such as healthcare. We propose an effective   new approach for  learning under extreme label noise,  based on under-trained deep ensembles. Each  ensemble member is trained with a subset of the training data, to acquire a general overview of the decision boundary separation, without focusing on potentially erroneous details. The accumulated knowledge of the ensemble is  combined to form new labels, that  determine a better class separation than the original labels. A new model is  trained with these labels to generalize reliably despite the label noise. We  focus on a healthcare setting and extensively evaluate our approach  on the task of sleep apnea detection. For comparison with related work, we additionally evaluate on the task of digit recognition. In our experiments, we observed performance improvement in accuracy from 6.7\% up-to 49.3\% for the task of digit classification and in kappa from 0.02 up-to 0.55 for the task of sleep apnea detection. 
\end{abstract}

\section{Introduction}
Recent advances in Machine Learning (ML) and sensor technology  have enabled the possibility of  reliable personalized healthcare diagnosis at home. For supervised learning,  proper labelling is needed for a classifier to generalize reliably.  However, labels can contain errors  due to human error, incomplete information, etc. This  poses a significant hindrance to proper generalization, which  can be very problematic in critical fields like healthcare since it can lead to inappropriate decision making with  potentially life-threatening consequences. Furthermore, such errors are  typically difficult and labor intensive to fix, since  labelling in  healthcare requires in most cases expert knowledge.

To address these issues we  develop a new  approach called  Evaluation Probabilities-based Distillation (EPD) based on recent key insights for memorization and predictive uncertainty in Deep Learning (DL) models \cite{lakshminarayanan2017simple,arpit2017closer}. EPD utilizes a DL-based ensemble   to create new labels  and perform Knowledge Distillation (KD) on a new DL model. A key characteristic of EPD is that it under-trains the members of the ensemble, i.e., for each member training stops prematurely, before the  loss is properly minimized. As such, EPD utilizes an  aggressive form of early stopping to achieve proper generalization, despite the label noise.  The main intuition  for this choice is straightforward: If a teacher  (i.e., the labelling) is partially erroneous but on a high-level correct, under-training can help  to avoid learning erroneous details of the teachers  beliefs. Furthermore,  training multiple models  and  using  different subsets of the data, is used to make the new labelling more robust  and  reduce  variance.  

In contrast to the majority of  related literature  which focuses on computer vision tasks, we focus on sensor time-series data. Specifically, we evaluate EPD for the task of  sleep apnea detection. Sleep apnea is a very common yet severely undiagnosed  sleep related breathing disorder which affects a large portion of the population. Sleep Apnea events are defined as the cessation of airflow for at least 10 seconds or reduced airflow by at least 30\% \cite{ApneaAASM}. To provide a basis for comparison with other works we additionally evaluate on common digit classification tasks (i.e.,  MNIST and SVHN datasets). For both domains, we explore commonly used artificial label corruption configurations and  more realistic settings to  measure the success and practical applicability of EPD. The contributions presented in this paper can be summarized  as follows: (1) We present a novel DL based  method for  learning under extreme noise conditions, (2) We showcase  state-of-the-art performance across multiple tasks and datasets, and (3) Through a variety of  experiments  we  evaluate the practical feasibility of the proposed approach, and empirically show that it constitutes a viable strategy for all investigated cases \footnote{We include relevant  proof-of-concept code at: https://github.com/konkoniknik/Noisy-Labels-EPD }.

\section{Method}

 In this section we  discuss  previous ideas that inspired EPD, highlight its novel aspects and describe in detail the proposed method.  Finally,  we discuss the derivation process, and the design choices of EPD.

\subsection{Inspirations and Characteristics}

 EPD is inspired by  three previous contributions.
First, the insight that DNNs learn real patterns first and then noise    \cite{arpit2017closer}. This is a consistent characteristic that potentially happens due to the more complex decision boundaries that stochastic label noise tends to create. We exploit   this characteristic, by under-training the members of the ensemble, to filter out  corrupt labels, through weak representation in the evaluation probabilities (EP, see next section). An obvious issue with this approach is that potential exceptions, that  otherwise would have been learned by fine tuning can  be easily missed by the under-trained classifiers. This leads us to the second basic inspiration, i.e, to use of ensembles of classifiers  as filters to clean noisy datasets from corrupted labels  \cite{brodley1999identifying}.  We  use a large committee of under-trained deep neural networks, called base models, trained on different subsets of the training data, and form the EP as out-of-bag predictions of these models. We do this   to give   classifiers   the opportunity to train with subsets that focus on different characteristics of the data. The goal is to enable different base models learn  specialized patterns. If this happens successfully, then we expect that  the correct class that corresponds to these patterns  would be represented in the EP.  In contrast to Brodley et al., we utilize random subsampling instead of cross-validation as we (1) want to have many different data combinations for the different models to train with, and (2) need many iterations per datapoint to form consistent EP. Additionally,  we label with the  EP all classes of each datapoint, instead of tagging a datapoint as correct or incorrect and then removing the incorrect datapoints. This   gives us more information for the decisions of the ensemble.  Third, we adopt the idea from  \cite{reed2014training}, to focus on states of the model in order to overcome label inconsistencies. However, instead of using the current state of the model (i.e., including an entropy term in our loss), we utilize many under-trained previous states  of our model. 

Based on the above, with EPD, we draw from the insights of these works in a novel manner and form a new labelling and a new weighing schema to enable a new classifier  reliably train despite extreme label noise.

\subsection{Method Description}
\label{Method}
We train multiple classifiers $h_i$. Each classifier $h_i,i=1...K$ is trained (under-trained)  on a random subset $D_{E_i}$ of a dataset  $D=\{(\mathbf{x}_i,y_i)\}_{i=1}^N$ where $\mathbf{x_i}$ are  datapoints and $y_i$   labels. $D$ is of size $N$, and all $D_{E_i},i=0..K$  of size $N_E$ with $N_E<N$. We  label the remaining data of $D$, i.e., $D-D_{E_i}$, with  $h_i$. This process is similar to an out-of-bag estimate, but we perform subsampling instead of bootstrap to train $h_i$. We repeat with all $K$ classifiers.

 For each datapoint $\mathbf{x}_j$ and for each class, we divide the number of classifiers that chose the class by  the number of classifiers that labeled   $\mathbf{x}_j$. Due to the  random sub-sampling process, each datapoint is labelled by a subset of the classifiers in the ensemble. Thus, we end-up with EP for $\mathbf{x}_j$ from the subset of the ensemble that labeled $\mathbf{x}_j$. By definition summing up all these class probabilities equals to 1.  Algorithm 1 summarizes the creation of the EP.

\begin{figure}[h]
\centering
  \includegraphics[scale=0.3]{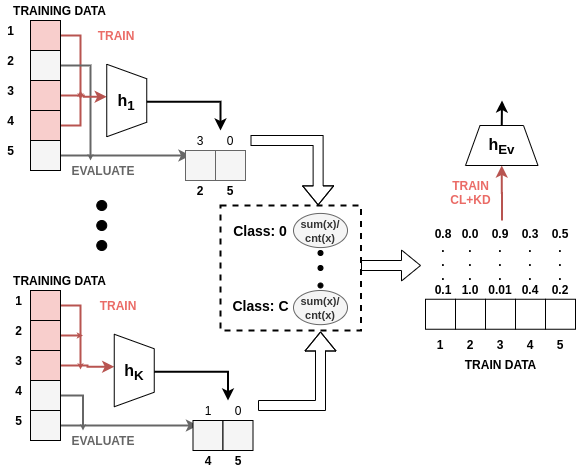}
\caption{ Structure of EPD. }
\label{fig:Method}
\end{figure}

\begin{algorithm}
\SetAlgoLined
\KwIn{Base Models: K, Dataset $D$, Subset size: $N_E$   }

Initialize  matrices $EP,S,C$ (Zeros)\;

\For{ i $<$K}{
 Initialize base model $h_i$\;
 
 $D_{E_i}$=$RandomSubsampling( D, N_E$)\;
 
 $h_i$=$Train(h_i$,$D_{E_i}$)\;
 
 \For{$\mathbf{x}_j \in (D-D_{E_i})$}{
 
 $S(\mathbf{x}_j)$= $S(\mathbf{x}_j)$+$h_i(\mathbf{x}_j)$\;
 
  $C(\mathbf{x}_j)$= $C(\mathbf{x}_j)$+1\;
 }
 }
 \For {$\mathbf{x}_j \in D$}{
  \eIf {$C(\mathbf{x}_j)==0$}{$Error$("More Models")}
{
  $EP(\mathbf{x}_j)$= $S(\mathbf{x}_j)/C(\mathbf{x}_j)$\;
  }
 }
 
 \KwOut{$EP$}
 \caption{Creation of Evaluation Probabilities}
\end{algorithm}
 
 Then we utilize the EP.  We train a new classifier $h_{Ev}$  with  the EP - or with a convex combination between the EP and the true labels-, as soft labels via   knowledge distillation \cite{hinton2015distilling}.  Additionally, for the training of $h_{Ev}$, we  define the weight of  $\mathbf{x}_j$ based on the maximum evaluation probability of $\mathbf{x}_j$ to rescale  the samples  based on the certainty of the ensemble (Figure \ref{fig:Method}).

\subsection{Methodological Derivation: Training $h_{Ev}$ with $EP$}
\label{sec:Requirements}

EPD is designed  to refine the ensemble filter  from \cite{brodley1999identifying}. Assuming a dataset $D$, the  empirical  risk  has normally the following form:

\begin{equation}
  \fontsize{8.5pt}{9.5pt} \selectfont
    \hat{R}(\mathbf{\theta})= \frac{1}{N}\sum_{(\mathbf{x}_i,y_i)\in D} L(h_{Ev}(\mathbf{x}_i,\mathbf{\theta}),\mathbf{y}_i)
\end{equation}

with $i=1...N$, and $\mathbf{y}_i=OneHot(y_i)$,  $L$ is a surrogate of the 0-1 loss and $\theta$ the parameter vector of a classifier $h_{Ev}$. However, after a filter like  the ensemble from \cite{brodley1999identifying} has been applied, the  empirical risk  is re-weighted with the following weighting scheme: The datapoints that fail the filter gain a priority of 0 and the ones that pass the filter gain equal priority. Formally, we define  the filter as a function $f:X\rightarrow \{0,1\}$ where $X$ is the input space,  0 corresponds to failure and 1 to success. Thus, we can partition $D$ into two subsets, i.e., comprising  datapoints that have failed ($S_0$)  and those that have succeded  ($S_1$) as such: $S_0=\{(\mathbf{x}_i,y_i)| f(\mathbf{x}_i)=0,(\mathbf{x}_i,y_i)\in D\}$ and $S_1=\{(\mathbf{x}_i,y_i)| f(\mathbf{x}_i)=1,(\mathbf{x}_i,y_i)\in D\}$. Now the empirical risk takes the following form:

\begin{equation}
  %\fontsize{6.5pt}{7.5pt} \selectfont
    \hat{R}_f(\mathbf{\theta})= \sum_{ (\mathbf{x_i},y_i)\in S_0} v_0 L(h_{Ev}(\mathbf{x}_i,\mathbf{\theta}),\mathbf{y}_i)+ 
    \sum_{ (\mathbf{x_i},y_i)\in S_1} v_1 L(h_{Ev}(\mathbf{x}_i,\mathbf{\theta}),\mathbf{y}_i)
\end{equation}

with $v_0=0$ and $v_1=\frac{1}{|S_1|}$. Because we utilize an ensemble of deep models that perform conditional density estimation,  an approximation of the conditional distribution for the training data is formed through the EP. Thus, we want to utilize the proposed method not only as filter but also to correct labels for which the ensemble exhibits a high probability that they are wrong. This probability is reflected in the EP. As such, we can generalize $f$ to identify  whether a certain condition is satisfied. Then for the data that satisfy this condition, we can  correct the label. We define:

\begin{equation}
  %\fontsize{6.7pt}{7.5pt} \selectfont
  f_{Ev}(\mathbf{x}_i) =\begin{cases}
      0  & \text{if $\max_c\{EP^{(c)}(\mathbf{x}_i)\}<T_0$} \\
   2 &\text{if $\max_c\{EP^{(c)}(\mathbf{x}_i)\}>T_2\cap arg\max_c\{EP^{(c)}(\mathbf{x}_i)\}\neq y_i$}\\
     1 & \text{otherwise}
  \end{cases}
  \label{Eq:filter}
\end{equation}

 where $EP(\mathbf{x}_i)$ corresponds to the EP vector for datapoint $\mathbf{x}_i$, and $EP^{(c)}(\mathbf{x}_i)$ corresponds the element of  $EP(\mathbf{x}_i)$ for class $c$. $f_{Ev}$ is a simple filter which filters data based on the confidence of the EP. We base our final empirical risk design on this filter. If $f_{Ev}$ is 0, then $\mathbf{x}_i$ is filtered-out. If  $f_{Ev}$ is 1 then $\mathbf{x}_i$ is not filtered-out. If $f_{Ev}$ is 2  then $\mathbf{x}_i$ is not filtered-out and we change the class of $\mathbf{x}_i$ to $\hat{y}_i= \arg\max_c EP^{(c)}(\textbf{x}_i)$. In Eq. \ref{Eq:filter}, we use $\mathbf{x}_i$ since $EP$ is defined only discretely for the datapoints of $D$. 
 In this case, we use as condition to change the label a threshold ($T_2$) for the maximum $EP^{(c)}(\mathbf{x}_i)$. This means that for $\mathbf{x}_i$,  we change $y_i$ to $\hat{y}_i$, if the maximum $EP^{(c)}(\mathbf{x}_i)$ exceeds $T_2$. Similarily, if the maximum $EP^{(c)}(\mathbf{x}_i)$ does not exceed $T_0$, we do not include the sample in the training data of $h_{Ev}$.  We  define  three non-intersecting sets $S_0,S_1$, and $S_2$, with $D=S_0\cap S_1\cap S_2$. Now, the empirical risk becomes:

\begin{equation}
  %\fontsize{7pt}{7.5pt} \selectfont
\begin{split}
     &\hat{R}_{fEv}(\mathbf{\theta})= \sum_{(\mathbf{x_i},y_i)\in S_0} v_0 L(h_{Ev}(\mathbf{x}_i,\mathbf{\theta}),\mathbf{y}_i)+\\
     & \sum_{ (\mathbf{x_i},y_i)\in S_1} v_1 L(h_{Ev}(\mathbf{x}_i,\mathbf{\theta}),\mathbf{y}_i)+ \sum_{ (\mathbf{x_i},y_i)\in S_2} v_2 L(h_{Ev}(\mathbf{x}_i,\mathbf{\theta}),\hat{\mathbf{y}}_i)
\end{split}
    \label{Eq3}
\end{equation}

with $\hat{\mathbf{y}}_i= OneHot(\hat{y}_i)$, $v_0=0$ and $v_1=v_2=\frac{1}{|S_1|+|S_2|}$. With this extension we can  allow corrections of labels as a pre-training step if certain conditions are being met. However, we identify two issues with this  design: (1) it could be difficult to properly tune parameters $T_0$ and $T_2$, and (2) it does not map directly the confidence of the ensemble  for a class of a datapoint onto the labelling and the weights of the weighting schema. Based on these requirements the following  two modifications  can be made:  First, the  maximum probability of the $EP$ of $\textbf{x}_i$ can be used as a weight of $\textbf{x}_i$ in the weighting schema of the empirical risk.  Second, a convex combination of the true label and the EP can be used instead of training directly with the true labels. In our case,  the maximum probability can again be used as the convex combination coefficient.  With these two modifications, the threshold parameters $T_1,T_2$ are removed, and  the confidence of the ensemble is directly  captured. Furthermore, the  characteristics of $\hat{R}_{fEv}$ are  preserved but with more focus on the EP than the real labels, which is desirable in an extreme noise setting. The final empirical  risk  takes the form:

\begin{equation}
  %\fontsize{7.5pt}{8.5pt} \selectfont
    \hat{R}_{Ev}(\mathbf{\theta})= \sum_{(\mathbf{x_i},y_i)\in D} v_i L(h_{Ev}(\mathbf{x}_i,\mathbf{\theta}),(1-\lambda_i)\mathbf{y}_i+\lambda_i EP(\mathbf{x}_i))
    \label{Eq4}
\end{equation}

with $v_i=\frac{max_c\{EP^{(c)}(\mathbf{x}_i)\}}{\sum_{j\in D} max_c\{EP^{(c)}(\mathbf{x}_j)\}}$, and  $\lambda_i=max_c\{EP^{(c)}(x_i)\}$. Thus, the original label will be mostly used and the sample will be de-prioritized during training, if  the ensemble is not confident for a sample. If  the ensemble is highly confident for a sample, the opposite will happen and regardless of label, the EP will dominate. Since the EP become more prominent and based on our empirical  investigation, we simplify this form and use directly EP as labels (i.e., in Eq. \ref{Eq4}, $\lambda_i=1$).  With this last change the use of the chosen weighting schema becomes less essential for the de-prioritization of erroneous labels. Instead, it acts as a scaling factor during training  for the  samples  which the ensemble is more confident of (i.e., less entropy).

%\subsection{TD: A few sentences about calibration}    

\section{Data and Experiment Description}

To evaluate EPD,  we use the following  principle throughout our experiments: we mainly  use datasets with very high generalization performance which we call clean datasets. Then we corrupt the clean training  datasets,  evaluate on the  clean test sets, and expect labels in the test set as the ground truth. This section describes the datasets and experiments using randomized label corruption as it is common in related literature, and experiments using "realistic" corruption to better reflect noisy labels in real world applications.

\subsection{Data }
In contrast to the main body of related literature, we focus on medical time series data. We use data from  Polysomnography (PSG) in sleep laboratory for OSA diagnosis. This data comprises a wide range of 
physiological signals, including the electrocardiogram (ECG), electroencephalogram (EEG), oxygen saturation (Sp02),
heart rate, blood pressure and respiration from the abdomen (RespA),
chest (RespC) and nose (NAF) etc. Since we want to capture a realistic application of EPD for a common medical condition, and explore its applicability in various domains, we focus  on respiratory sensor data. We mainly use the \textbf{Apnea-ECG} \cite{goldberger2000physiobank, penzel2000apnea},  ($AE$)  which is an open PSG dataset from Physionet, containing  RespA, RespC, NAF, SpO2 and ECG sensor data. $AE$ has been used in the Computers in Cardiology challenge  and  contains high quality data. In $AE$ 8 recordings  contain data from all  sensors. Each recording has a duration of 7-10 hours. Labels are given for every one-minute window of breathing, and identify which minutes are apneic and which are not. We   aim to capture the behavior of EPD for data gathered from all different sensor devices that focus on respiratory signals. Thus,  we evaluate on  signals  gathered with the use of a respiratory belt on the abdomen of the patient (RespA:$AE$), with the use of a nasal thermistor from the nose (NAF:$AE$), and with the use of a pulse oxymeter that measures oxygen saturation in the blood (Sp02:$AE$). We choose RespA over RespC as the belt sensor  due to its better  behavior in our preliminary evaluation. Furthermore, the lower quality \textbf{MIT-BIH} \cite{ichimaru1999development}, ($MB$) dataset is used to corrupt $AE$. 
Finally, we use two well-established datasets from digit classification,  i.e., \textbf{MNIST}  \cite{lecun1998gradient}, ($M$)  and \textbf{SVHN} \cite{netzer2011reading},($S$) to provide a basis for comparison   with related literature and to investigate the generalizability of EPD on different tasks. We discuss the datasets in more detail the Technical Appendix.

\subsection{Experiment Description}
\label{sec:ExpDescr}

The goal of the experiments is to quantify the impact that EPD has on classification performance. Since  a noisy baseline makes evaluation non-trivial  due to the lack of reliable ground-truth, it is common practice to use a noise-free dataset and add randomized noise to corrupt the training labels. While this can guarantee correct evaluation results, it is a strategy which in many cases is unrealistic. Therefore, we perform two types of experiments.

\begin{figure}[h]
\begin{subfigure}{0.325\columnwidth}
\includegraphics[width=\columnwidth]{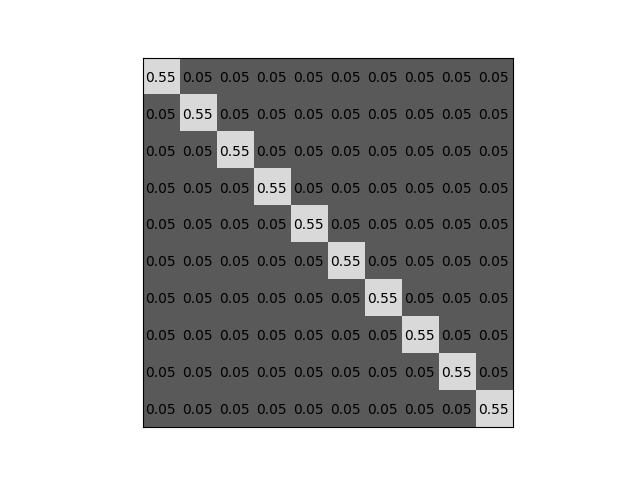}
\caption{Symmetry-0.45} \label{fig:1a}
\end{subfigure}
\begin{subfigure}{0.325\columnwidth}
\includegraphics[width=\columnwidth]{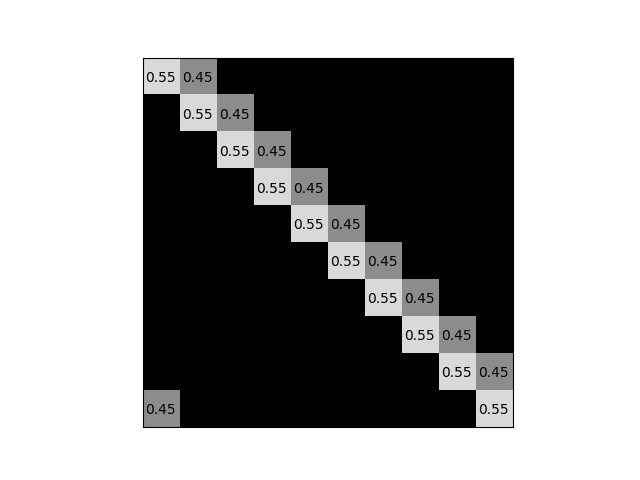}
\caption{Pair-0.45} \label{fig:1b}
\end{subfigure}
\begin{subfigure}{0.325\columnwidth}
\includegraphics[width=\columnwidth]{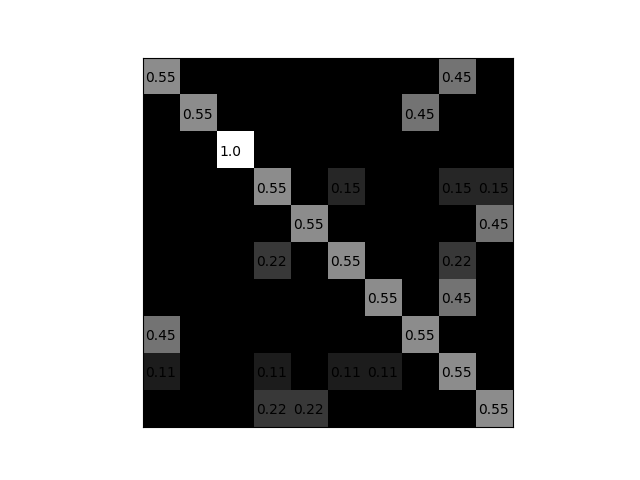}
\caption{Alphanum.-0.45} \label{fig:1c}
\end{subfigure}
\caption{ Examples of the noise transition matrix for  10 classes. The Symm. type gives equal probability for all  non-true classes, (a). The Pair type is similar to symmetry, but all the probability to change the class is  put in one other class, (b). In this example as well as  in the experiments, we use the next class to the right. In (c), change is done based on common misidentification patterns. } \label{fig:NoiseTrMatr}
\end{figure}

For the first experiment type, we use the clean   datasets (i.e., $AE$,$M$ and $S$) and corrupt them using artificial randomized noise sources. Then  we train with the corrupted training data, and evaluate on the clean test data, to get an indication of how well  EPD can generalize despite the noise. To corrupt the datasets we use of a noise transition matrix to change the class $i$ of a sample to another class $j$ with the probability mapped in the respective $(i,j)$ element of the matrix. We experiment with the two types of transition matrices commonly used in literature, i.e., pair and symmetry (shown in  Figures \ref{fig:1a} and \ref{fig:1b}).  For the second experiment type, we perform different experiments where we corrupt the datasets in more realistic and task-specific ways than in the previous case. For the digit classification task, we create a noise transition matrix based on common human Alphanumeric misidentification patterns indicated by relevant medical literature \cite{MisDIDigit}. We show the noise transition matrix  for this task in Figure \ref{fig:1c}. Then, for the apnea dataset $AE$ we: (1) introduce  realistic  label noise and a domain shift by adding $MB$ data into the $AE$ training data and (2)  investigate whether the proposed approach can be used  to enhance the labelling of apnea data  of a non-expert.

\section{Evaluation}

In this section we describe the preprocessing steps we performed, the  decisions we made regarding general and method-specific hyper-parameter values, and the experiments to evaluate the proposed approach. As baseline, we use in all experiments the results gained with the $h_{Ev}$ model, trained with equal weights per datapoint and with the true labels (i.e., no KD) regardless of noise level.

\subsection{Preprocessing,   Metrics and Hyper-Parameters}

In all experiments, we execute the same preprocessing steps for the datasets. We rescale the pixel values of all samples of the digit datasets $S$ and $M$ from 0-255 to 0-1. Additionally, for $S$ we restructure the data in similar form to $M$ (from $32\times32\times3$ to $28\times28\times1$). Furthermore, both digit datasets have a predefined test sets which we will use.

In both sleep apnea datasets the data is standardized per physiological signal, downsampled to 1Hz,  shuffled randomly and divided into train and test sets. We perform the last step since no predefined test sets exist in the respiratory datasets.  The percentage of data used as test set changes depending on the experiment. Furthermore, the labels in $MB$ are given every 30 seconds and in  $AE$   every 60 seconds. Therefore, we adapt  the labelling in $MB$ to 60 seconds with the following rule: if both 30 second labels are non-apneic then the 60 seconds label is  non-apneic, elsewise it is apneic. 

We evaluate   the  classification performance on the clean test set. For the Digit Classification experiments we use Accuracy as performance metric. For  Apnea Detection, which is a two class problem,  we use the Kappa statistic \cite{cohen1960coefficient}, since Kappa   takes into account  random agreement. This makes it a suitable choice for the task of apnea detection. Additionally, for  all results the standard error of the mean is presented.

 In all experiments we use convolutional neural networks  with a batch size of 128, and a  learning rate of 0.001. For brevity, we include the architectures used in the Technical Appendix.  For the evaluation we follow the  conventions from \cite{han2018co}, i.e., in all experiments the baseline model and $h_{Ev}$ are trained for 200 epochs, and  the  generalization performance is calculated from the average performance of the last 10 epochs.

Furthermore, $K=100$ base models  are used for EPD,    1/3 of the training dataset is used  as evaluation dataset, and 2/3 for the actual training. In most experiments, the base models are trained  for  4 epochs. The choice of the amount of epochs to train the base models is not trivial  since the use of a noisy validation set to choose the best performing epoch does not guarantee good generalization performance on the clean test set.  Empirical investigation showed that a more reliable metric is the minimum mean entropy of the EP. Furthermore, this metric is in-line with the under-training principle. The minimum EP entropy is  reached in most configurations early during training, and close to the epoch where the maximum generalization performance is reached on the clean test set.  We discuss in more detail  the reasoning followed to choose the values of these hyper-parameters in the Technical Appendix.

\subsection{Randomized Corruption}

We use the Pair and Symmetry noise transition  configurations and investigate the performance of EPD in comparison to the baseline, for various noise levels.

\subsubsection{Digit Classification}

Table \ref{table:Digit_main} shows the results for the $S$ and $M$ datasets under the Pair and Symmetry label corruption configurations. As expected  EPD outperforms Baseline training in all cases by a wide margin. Notice that  the pair noise configuration is harder than the symmetry since each class can only change to the one other class. As a result this class will be overly represented in place of the original class, making the training inherently harder, especially as the corruption probability approaches 50\%.

  \begin{table*}[h]
  \fontsize{9pt}{10pt} \selectfont
\centering
\begin{tabular}{ p{1.3cm}|p{1.65cm}|p{1.65cm}||p{1.65cm}|p{1.65cm}|p{1.65cm}}
%\multicolumn{6}{|c|}{$S$ and $M$  Accuracy under Noise (Uncorrupted Performance: $M$:TD,$S$:TD) } \\
%\hline
Method& Pair-0.40 & Pair-0.45 & Sym.-0.20 & Sym.-0.50 & Sym.-0.63  \\
\hline
S:Baseline&0.587$\pm$0.001&0.523$\pm$0.002&0.835$\pm$0.001&0.593$\pm$0.002&0.429$\pm$0.002\\
S:EPD&0.849$\pm$0.002&0.736$\pm$0.005&0.905$\pm$0.001&0.877$\pm$0.001&0.85$\pm$0.001\\

\hline
M:Baseline&0.621$\pm$0.002&0.561$\pm$0.002&0.914$\pm$0.001&0.639$\pm$0.002&0.473$\pm$0.001\\
M:EPD&0.979$\pm$0.001&0.95$\pm$0.002&0.981$\pm$0.001&0.973$\pm$0.001&0.966$\pm$0.001\\
\end{tabular}
\caption{$S$ and $M$ performances  (accuracy) of EPD for various noise levels. Clean performances: $M$:0.991, $S$:0.920 }
\label{table:Digit_main}
\end{table*}

\begin{figure}[h]
\begin{subfigure}{0.49\columnwidth}
\includegraphics[width=\columnwidth]{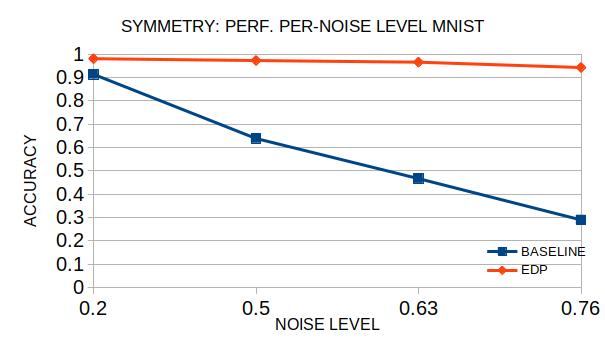}
\caption{} \label{fig:2a}
\end{subfigure}
\hspace*{\fill}
\begin{subfigure}{0.49\columnwidth}
\includegraphics[width=\columnwidth]{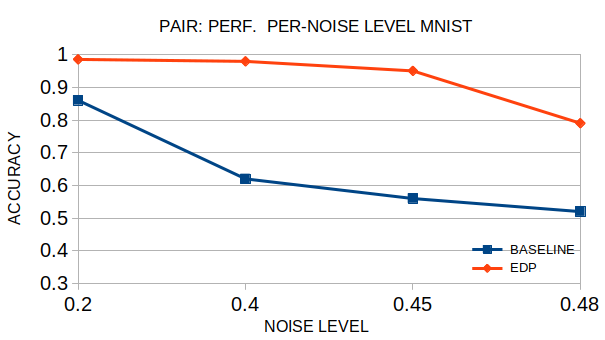}
\caption{} \label{fig:2b}
\end{subfigure}

\begin{subfigure}{0.49\columnwidth}
\includegraphics[width=\columnwidth]{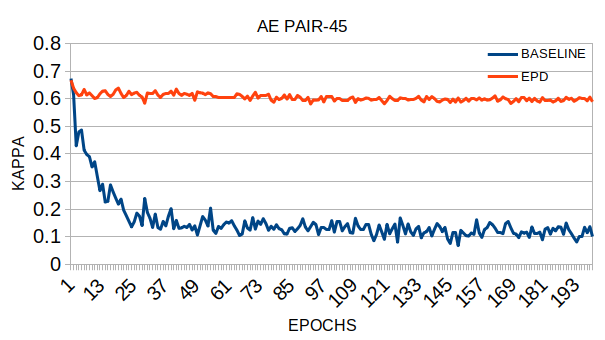}
\caption{} \label{fig:4b}
\end{subfigure}
\hspace*{\fill}
\begin{subfigure}{0.49\columnwidth}
\includegraphics[width=\columnwidth]{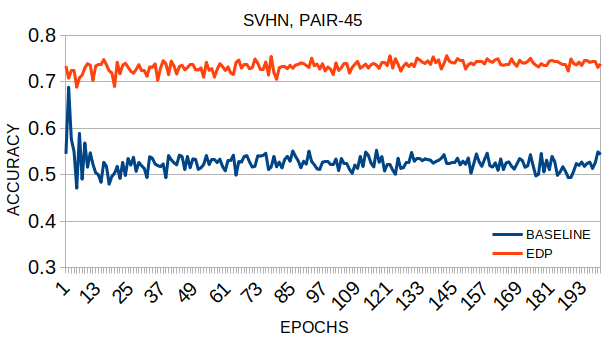}
\caption{} \label{fig:3b}
\end{subfigure}
\caption{ Results from experiments with $M,S,NAF:AE$. (a) and (b) show the resilience of EPD and Baseline  against  Sym. and Pair  noise corruption, in each case for varying degrees of noise for $M$.   (c) and (d) show performance through 200 epochs for Baseline  and EPD under Pair-0.45 Noise configurations for $S$ and $NAF:AE$. } 
\label{fig:MNIST_Res}
\end{figure}

Figure \ref{fig:MNIST_Res} shows several performance characteristics for the two datasets. In Figure \ref{fig:MNIST_Res},(a),(b)  we  notice for $M$,  for the different corruption configurations and noise levels that  with decreasing true class representation, the performance of the baseline decreases steadily. In contrast,  EPD manages to perform well even in cases of extreme noise,  (e.g., the cases of Symmetry-0.76 or Pair-0.48). 
In the second row, (Figure \ref{fig:MNIST_Res}, (d)) we show the model's performance throughout training for the baseline in comparison to EPD for $S$. The initial performance of Baseline is high, but it quickly drops, as expected. Contrary to that, $h_{Ev}$, which is trained with EPD is mostly not affected from the label noise. We include additional similar graphs for other noise configurations  in the Technical Appendix.

 Though EPD is  inherently a simple method, we achieve state-of-the-art results  compared to other well-established methods like \cite{han2018co} ($M$) or \cite{lee2019robust} (S). We expect that  the success of EPD depends on: (a) the ability of the base classifiers to prioritize learning clean samples during learning, i.e., under-training is successful, (b) on the ability of the ensemble to compensate from erroneous beliefs on certain datapoints. 

\subsubsection{Sleep Apnea Detection}

 For the sleep apnea detection experiments, we investigate the proposed approach  with  
three   signals from the $AE$ dataset (i.e., Resp A, NAF and Sp02). Similarly as before we corrupt the labels with random noise based on the noise transition matrix. However, in this case,  we only have two classes, i.e., apneic and non-apneic minutes. Thus, we practically  investigate  the Pair noise type.

Table \ref{table:AE_main} shows the comparative results between Baseline training and EPD across the three respiratory sensors. In all cases, we notice a steep drop in performance for the Baseline  with increasing noise. The performance drop is substantially smaller for  EPD. Note that the widest margins in performance difference between Baseline and EPD is for the NAF signal. Figure \ref{fig:MNIST_Res} (c) shows the performance (kappa) for the NAF signal, throughout training for noise level of 45 percent (Pair-45, average of 3 runs). We observe similar characteristics as before. The Baseline  has a steep drop early-on in the training process. This is not the case for  EPD, which is only initially affected by the noise level,  and not throughout the whole training process.

 \textbf{Significance for  Condition Detection:} From the results we notice that the performance for  all signals is  significantly affected from the noisy labelling, especially for larger noise levels.This means that a "careless" annotator would affect negatively the correct identification for this medical condition. However, the use of the proposed method is highly likely to allow  an expert  have a much wider error margin during labelling. Depending on the goal, this margin can be translated as the relaxation of the constraints placed upon the quality of the annotator's expertise.

 \begin{table}[h]
   %\fontsize{9pt}{10pt} \selectfont
\centering
\begin{tabular}{ p{1.9cm}|p{1.6cm}|p{1.6cm}|p{1.6cm}}
%\multicolumn{4}{|c|}{$AE$ Kappa under Noise.  Clean performances NAF:TD, RespA: TD,SpO2:TD  } \\
%\hline
Noise& Pair-0.20 & Pair-0.35 & Pair-0.45 \\
\hline
NAF:Baseline&0.581$\pm$0.004&0.294$\pm$0.005&0.103$\pm$0.008\\
NAF:EPD&0.888$\pm$0.003&0.829$\pm$0.005&0.584$\pm$0.020\\
RespA:Baseline&0.568$\pm$0.007&0.267$\pm$0.007&0.098$\pm$0.010\\
RespA:EPD&0.851$\pm$0.003&0.813$\pm$0.007&0.547$\pm$0.020\\
Sp02:Baseline&0.660$\pm$0.007&0.425$\pm$0.006&0.179$\pm$0.013\\
Sp02:EPD&0.828$\pm$0.004&0.811$\pm$0.003&0.644$\pm$0.025\\

\end{tabular}
\caption{ $AE$  perf. (kappa) of EPD for various noise levels. Clean perf.: NAF:0.919, RespA:0.895, SpO2:0.875 }
\label{table:AE_main}
\end{table}

\subsection{Realistic Corruption}
In this section we evaluate EPD on realistic sources of noise.
\subsubsection{Digit Classification: Alphanumeric Misidentification Patterns}

In medical literature \cite{MisDIDigit}, common alphanumeric misidentification patterns are identified. We use these patterns to create a noise transition matrix (Figure \ref{fig:NoiseTrMatr} (c)) to corrupt the true labels in a realistic manner.  We use noise level of 0.45, since it is a representative  level of an extremely noisy setting. Furthermore, for some classes we are not able to exceed noise level of  0.5 without the noise overtaking the true class. We assign this probability mass equally to the classes that can potentially be swapped in place of the true class. For example, in the case of digit 3, which label can be changed to 5,8 or 9, we assign equal probability of 0.15 to these three  swaps. We call this  type of corruption Alphanumeric-0.45. We execute the experiment for $S$ and $M$ datasets. We repeat the experiments 10 times.

For $M$ the average accuracy of Baseline is: 0.633$\pm$0.002, and with EPD: 0.949$\pm$0.001. For $S$: Baseline: 0.612$\pm$0.001, EPD: 0.776$\pm$0.002. As expected,  EPD  significantly outperforms  Baseline in both datasets.

\subsubsection{Apnea Detection: Corrupting Clean Dataset with Noisy Dataset}

We combine part of $AE$  and $MB$  to create a joint training dataset. $MB$ is noisier than  $AE$  in terms of data and labelling. Thus, we corrupt the clean training datapoints of $AE$ with the noisier  $MB$ datapoints. We use  a test set comprised only with datapoints from $AE$ (we use 75\% of $AE$). We use the NAF signal  as  it is the most common signal   of $MB$.  We investigate different percentages of inclusion  of the remaining  25\% of $AE$ in the new joint training dataset (i.e., we include 100\% of $MB$ and either 25\%,12.5\%,5\%,1.25\%,or 0\% of $AE$). Less inclusion of $AE$ data in the Joint training  dataset corresponds to "noisier" labels in the training data. We repeat the experiment 20 times.

\begin{figure}[h]
\begin{subfigure}{0.49\columnwidth}
\includegraphics[width=\columnwidth]{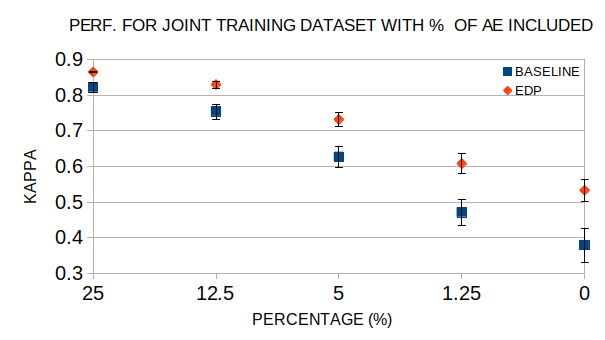}
\caption{} \label{fig:5a}
\end{subfigure}
\hspace*{\fill}
\begin{subfigure}{0.49\columnwidth}
\includegraphics[width=\columnwidth]{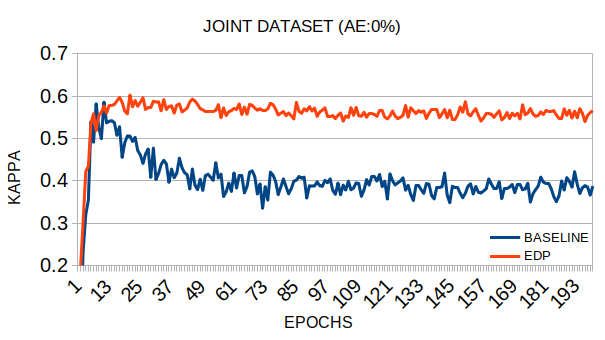}
\caption{} \label{fig:5b}
\end{subfigure}
\caption{ (a): Performance for different percentages of $AE$  in Joint dataset.  (b): Performance, $AE$ inclusion 0\% } 
\label{fig:Joint_Res}
\end{figure}

  In all cases,  EPD is superior to Baseline training, as expected (Figure \ref{fig:Joint_Res} (a)). As smaller percentages of $AE$ are included in   the Joint training data, we expect drop in performance  due to the noisy labelling and the domain shift imposed in the input space. Thus, we expect both cases to drop in performance as we include smaller parts of $AE$ . Indeed, we observe such a behavior from Figure \ref{fig:Joint_Res} (a). Notice also that the difference  between Baseline and EPD increases as the percentage of $AE$ decreases.  In Figure \ref{fig:Joint_Res} (b), we show the test performance  during training, for  $AE$ inclusion of 0\%. We notice a similar behavior as in the previous sections.

 \subsubsection{Apnea Detection: Non-Expert Labelling}

 A non-expert  labelled a small subset of $AE$ (corresponding to 5\% of the whole data), for the three sensors we utilize in this work i.e.,Resp A, NAF, and Sp02. To do this, a "hint" of 10 randomly sampled datapoints was presented to the non-expert together with their true labels, to help  the non-expert learn  patterns per-class. Then, the non-expert proceeded to the actual labelling of a random part of $AE$ (5\% of $AE$).  We utilize this dataset  and  train a model  with these data and labels. We evaluate the generalization performance  of the model with the other 95\% of $AE$ dataset. We repeat this procedure for all three sensors.
 In this experiment, the majority of conventions and configurations are as the ones described above. However, since we have a very small training dataset, we need to keep per base classifier $h_i$ as much of the data as possible for training. For this reason we use 1/10 of the training data as evaluation datasets instead of 1/3. Furthermore, to partially compensate for this and  create reliable EP, we use more base models ($K=200$) than in the default configuration. Finally, based on the EP-entropy criterion (see Technical Appendix) used to determine how much to train the base classifiers, we notice that 4 epochs are insufficient for such a small dataset. Thus, we train for 20 epochs the base models. Again, we compare the baseline to EPD. Given the non-expert labels, we repeat the experiment (i.e., training and evaluating on the noise-free test set) 10 times.
  
 The agreement of the non-expert labelling  with the true labels (i.e., the expert's labelling) for the three sensors is (Accuracy): Resp A: 0.855, NAF: 0.94, Sp02: 0.745. The generalization performance on the test set (Kappa) is: Resp A: Baseline:  0.730$\pm$ 0.003, EPD: 0.788$\pm$0.001; NAF: Baseline: 0.838$\pm$0.004, EPD: 0.860$\pm$0.004; Sp02: Baseline: 0.588$\pm$0.008, EPD: 0.743$\pm$0.005. In all cases EPD outperforms the baseline. The results are statistically significant  based on the one-tailed  t-test for all sensors. A clear correlation between the performance difference of Baseline and EPD on the one hand, and  the agreement between labellings on the other hand can be observed.
 
\subsection{Discussion}

From the above experiments, we identify that EPD offers a reliable approach for training under  conditions of extreme label noise, even  in realistic settings as long as proper parametrization conditions are being met. The results  of our experiments are  promising, and showcase  the versatility of EPD even when the strict definition of  noise is laxed.

An important empirical observation is that, as  expected, EPD cannot provide a reliable training regime in cases where the noise levels are so high that they form a  bias towards a wrong class, i.e., where, for certain regions of the feature space a consistent wrong class belief is formed by an erroneous labelling strategy. In such cases, where the majority of labels leans on one wrong class in such a way that it can be wrongfully learned, the under-training could potentially have a negative effect in the learning of the correct decision boundaries. However,  these cases correspond to principle  mistakes of the labelling strategy, and not in occasional mistakes that constitute noise in the labels. Examples of such cases in the medical domain could be that of an untrained annotator which has misunderstood basic identifiers of a condition, and makes repeatedly the same errors, or a rule-based software that is not properly tuned for a particular individual.   We hypothesize that for such cases a noise correction learning  framework would not be the better choice, and different bias correction strategies would need to be implemented to help a new model to generalize correctly.  However, for  the problem EPD is designed to tackle it yields promising and reliable empirical results in a variety of  tasks and domains.

\section{Related work}
%Auta ta thelw alla den kserw akoma pou na ta balw.
%1) In \cite{tumer1996estimating}, Tumer et al. provide a way to get an estimate of the Bayes Error with the use of classifier combination. In two toy-datasets they achieve superior estimation than other techniques like the Mahalanobis bound.(Discussion not sure its for RW)\\
%7)  Zhang et al. \cite{zhang2016understanding}, showcase that deep networks with large  parameter number in relation to the sample size can perfectly learn the training labels, even if they are corrupted. More generally, they showcase that deep network, can easily learn random labels. 

There exist a large body of  literature addressing  corrupt labels, which is well  captured in \cite{karimi2019deep}. From this body of work, we present only works that have key similarities to our own.  We separate this section based on two types of works: (1) works that use multi-network frameworks or KD (2) works that use data weighing strategies.

Many works utilize KD or multi-network configurations to achieve better training under label noise. Examples include \cite{reed2014training,zhong2019unequal}, which utilize the current state of the model and update the prediction labels with the current model prediction . Other  approaches that utilize a student teacher configuration include \cite{yu2019uncertainty,li2017learning}. \cite{li2017learning}, propose the use of KD from an auxiliary model trained on a small clean dataset to guide the training of  a model  trained on a larger noisy dataset using a convex combination of the noisy labels and the output of  the auxiliary model. We experimented with  a convex combination of the true label and EP. However,  our experiments show that  the results with the convex combination were inferior to those with only  EP.   \cite{malach2017decoupling}  train two networks which are only updated  when they disagree for a given instance.    \cite{lakshminarayanan2017simple} propose the use of ensembles of DNNs to quantify  predictive uncertainty. The insights obtained from this work are used together with MC-dropout \cite{gal2016uncertainty} by  \cite{kohler2019uncertainty} to  detect noisy labels. They propose the use of uncertainty estimates to identify a good epoch for relabelling a percentage of the lowest confidence data.   Contrary to this work, we base our decision of how much to train on the entropy of the EP. Furthermore, another key difference between EPD and the above approach is that in EPD the EP are formed as out-of-bag estimates from the different base classifiers of the ensemble. As such, they represent a form of pseudo-generalization. None of the base classifiers contributing to the EP for a  datapoint $\textbf{x}_i$ has made any association about its label beforehand.  Finally, \cite{ostyakov2018label} create out-of-fold logits based on  the predictions of base classifiers with the use of a base-ensemble of heterogeneous classifiers and  LBGM \cite{ke2017lightgbm}. Despite the similarities with this approach, we utilize under-training as a way to more efficiently identify noisy labels and exploit the out-of-fold predictions (EP)  as a form of curriculum for the training process. Furthermore, we  use the EP to identify the proper amount of training for the ensemble's base models.

Regarding data re-weighing strategies,   \cite{shen2018learning}   observe that training accuracy is evolving differently for noisy and clean samples. Based on this observation they propose  alternating between selecting a subset of the training samples that has the smallest loss, and training with this sample. Other similar approaches include \cite{xue2019robust,shu2019tug,ding2018semi}.\cite{ding2018semi} identify incorrect labels with the use of a DNN. Labels that differ from the predictions of the DNN  (trained on the noisy dataset) are removed. The remaining data are sorted based on prediction confidence and labels below a certain confidence threshold are also removed. Then  semi-supervised learning is applied to train with both data that contain and do not contain labels.
A different approach  partially based on data reweighing  is  co-teaching  \cite{han2018co}. Han et al. utilize  two networks, and each network learns with the subset of the extracted mini-batch, which contains  the instances  with the smallest loss of the other network.  \cite{yu2019does} expand upon co-teaching by applying the same procedure, but focusing on a subset of the data, which contains only points for which the two networks disagree.  \cite{jiang2017mentornet}  introduce MentorNet, which learns a curriculum to weigh training samples' priority during the training of a student network. A key difference  between   the above approaches and EPD stems from the fact that the weighting schema in EPD originates from the accumulative knowledge of a deep ensemble. However, in EPD the reweighing  plays a secondary role due to the use of KD, and it is only used to de-prioritize uncertain samples.

\section{Conclusions and Future Work}

In this work we investigated a simple, yet effective approach for successful learning under extreme label noise. Contrary to related literature, EPD re-uses the cumulative knowledge of the ensemble back in the training data in such a way that it takes advantage of the generalization capabilities of independent parts of the ensemble and forms new labels for the training data. With the use of KD,  a new model is trained with the new labels. As such, the ensemble is used  only for the synthesis of the new labels, and only the new model needs to be used   after the procedure is finished.

 We empirically show the success of EPD in a variety of settings, configurations and noise levels, for two  real world tasks. We observe that EPD can offer a general and reliable solution to the problem of learning under label noise. Our results serve as indicators of the applicability of EPD towards more reliable condition diagnosis. Labelling assistance can be another interesting  application direction.

A natural next step is the identification of more consistent errors in regions of the feature space which are based on wrong beliefs of the annotator and not random mistakes. We hypothesize that an  approach which incorporates elements from unsupervised learning  could  potentially be able to assist towards the identification of these wrongful biases. Based on this, we want to incorporate such elements to the EPD framework and provide a more general  solution.

\bibliographystyle{unsrt}
\bibliography{mybibfile}

\section{Ethics Statement}

In the context of sleep apnea detection, noisy or inconsistent labelling is a real problem. Even trained sleep experts disagree to a certain extent about apneic and non-apneic events. Inconsistent  labelling can lead to wrong conclusions with respect to treatment. Thus, reliable classification for such tasks can have a positive impact on well-being and reduce costs in the health sector. The primary motivator for the approach presented in this work is such a healthcare-oriented setting. However, since the approach is general, it can also be used in other settings. 

%\clearpage
\appendix

\section{Appendix A: Datasets}
In this Appendix we describe in more detail the datasets used in our experiments:

\begin{itemize}
\item \textbf{MNIST} ($M$)  \cite{lecun1998gradient} is a well-known  dataset which contains 60000 28$\times$28  images of digits (handwritten black and white images of 0-9) as a training set.  The test set is composed of 10000 images.
\item \textbf{SVHN} ($S$)\cite{netzer2011reading}  is  a real-world image dataset. $S$ is   obtained from house numbers in Google Street View images.  It contains 73257 digits (32$\times $ 32 colored images) for training, 26032 digits for testing, and 531131  less difficult samples that can be used as additional training data. We use only the original training dataset of 73257 digits.
\item \textbf{Apnea-ECG} ($AE$)\cite{goldberger2000physiobank, penzel2000apnea}  is an open PSG dataset from Physionet, containing  RespA, RespC, NAF, SpO2 and ECG sensor data. $AE$ has been used in the Computers in Cardiology challenge \cite{penzel2000apnea} and  contains high quality data. From the 35  recordings in the dataset, 8 recordings (namely a01-4,b01,c01-3) contain data from all  sensors. Each recording has a duration of 7-10 hours. The sampling frequency  of all sensors is 100Hz, and labels are given for every one-minute window of breathing. The labels identify which minutes are apneic and which are not (i.e., if a person experiences an apneic event during this minute). In this work, we   aim to capture the behavior of EPD for data gathered from all different sensor devices that focus on respiratory signals. Thus,  we evaluate on respiratory signals  gathered with the use of a respiratory belt on the abdomen of the patient (RespA:$AE$),  with the use of a nasal thermistor from the nose (NAF:$AE$), and with the use of a pulse oxymeter that measures oxygen saturation in the blood of the patient (Sp02:$AE$).

 \item \textbf{MIT-BIH} ($MB$) \cite{ichimaru1999development}  is an open dataset containing  recordings from 18 patients, with different  respiratory sensor signals.  In 15 recordings, the respiratory signal  has been collected  with NAF. Since this is the signal with the most recordings in $MB$, we use it for our experiments. Due to   misalignment of the different signals and lack of labels for the apneic class in 4 recordings, we utilize 11 of the 15 recordings (slp60,slp41 and slp45 and slp67x are  excluded). Related literature \cite{kristiansen2018data} suggests that the data/labelling quality of the $MB$ dataset is  low. This leads to low  classification performance for $MB$ compared to $AE$. This characteristic of $MB$ is important for our experiments.  The labels are given for every 30 seconds and the sampling frequency of all sensors is 250Hz.
 \end{itemize}

\begin{table*}[h!]
\centering
\begin{tabular}{ | p{2.2cm}| p{2.6cm}| p{2.8cm}|p{1.7cm}|}
\hline
\multicolumn{4}{|c|}{Architectures} \\
\hline
 & MNIST & SVHN & Apnea\\
\hline
Conv,(MP,BN)&$1 \times 8\times 5\times5$&$1 \times 32\times 5\times5$& $1\times 16 \times 5$ \\
Conv,(MP,BN)&$8\times 16\times5\times5$&$32\times 64\times 3\times 3$&$ 16\times 32\times 5 $\\
Conv,(MP,BN)& $16\times 32\times5\times5$  &$64\times 128\times5\times5$ &$ 32\times64\times 5$ \\
FC,D,(BN)&$(7\times 7 \times 32)\times 512$ &$(7\times 7\times128)\times 3072$ &$ (8\times64)\times 64$\\
FC,D,(BN)&$512\times 256$ &$3072\times 1024$&$ 64\times 32$ \\
out&  $256\times 10$&$1024\times 10$& $32\times2$   \\
\hline
\end{tabular}
\caption{ Used  Architectures.  In  all  experiments  for  each dataset   we   use   the   defined   architecture   (Conv:   input channels$\times$output  channels$\times$filter,  MP:  Max  Pooling-not used in Apnea datasets-, fc: Fully connected, input$\times$output, BN: Batch Normalization -used only on SVHN, D:Dropout-0.75).}
\label{table:Archs}
\end{table*}
\section{Appendix B: EPD Hyper-Parameter Tuning}

In this Appendix, we discuss the hyper-parameter tuning for EPD.

\subsubsection{Epoch Choice for Base Classifiers}

\begin{figure}[h]
\begin{subfigure}{0.49\columnwidth}
\includegraphics[width=\columnwidth,,height=2.5cm]{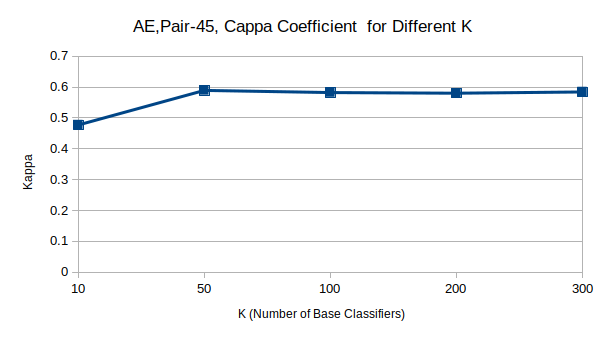}
\caption{} \label{fig:6a}
\end{subfigure}
\hspace*{\fill}
\begin{subfigure}{0.49\columnwidth}
\includegraphics[width=\columnwidth,height=2.5cm]{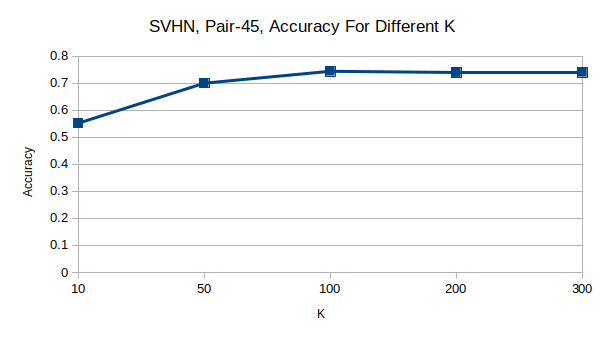}
\caption{} \label{fig:6b}
\end{subfigure}
\hspace*{\fill}

\begin{subfigure}{0.49\columnwidth}
\includegraphics[width=\columnwidth,,height=2.5cm]{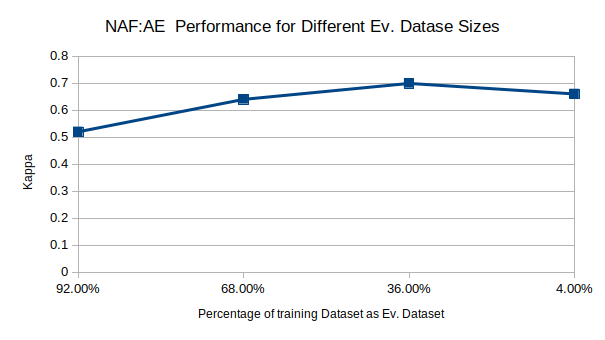}
\caption{} \label{fig:6c}
\end{subfigure}
\begin{subfigure}{0.49\columnwidth}
\includegraphics[width=\columnwidth,,height=2.5cm]{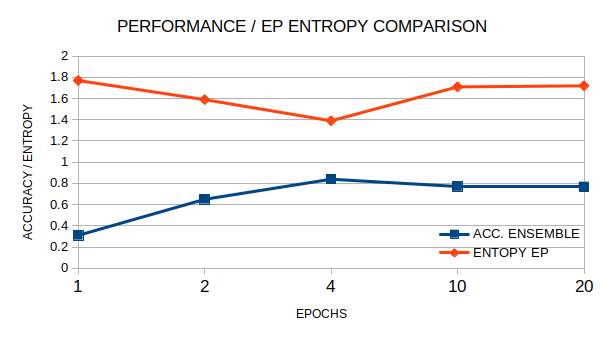}
\caption{} \label{fig:6d}
\end{subfigure}
\caption{  Performance for $S$ and $AE$ datasets for different values of $K$ (Fig. (\ref{fig:6a}), (\ref{fig:6b})).   Performance for EPD NAF:$AE$, for different Evaluation dataset sizes (Fig. (\ref{fig:6c})). Comparison of performance to EP-Entropy for $S$ Symmetry-0.37  (Fig. (\ref{fig:6d})).} 
\label{fig:Discussion_K}
\end{figure}

One of the most important parameters of EPD is the number of epochs used to train the base classifiers $N$. We base our choice on the insights from \cite{arpit2017closer}, and we choose to under-train the networks, so that we avoid fitting on the noisy labels. We notice that the entropy of the EP is a relatively reliable empirical  measure of the generalization capabilities of the ensemble, which directly impacts  the generalization capabilities of $h_{Ev}$. We hypothesize that the use of different evaluation datasets for the different base classifiers  as well as the use of generalization for the specification of the EP have a beneficial effect towards appropriate calibration of the EP of the different samples. Based on this hypothesis we use the mean entropy of the EP as an estimate of the true correctness likelihood of the dataset. Thus, we choose the number of epochs  $N$ with which we train the base classifiers based on the minimum mean EP entropy reached among a set of different potential values for $N$. Figure (\ref{fig:6d}) presents the performance-to-EP-Entropy comparison for $S$ (Orange-Blue respectively). We notice a clear correlation between the two measures. Similar associations were made across the  majority of the different noise configurations and datasets.

\subsubsection{ Number of Base Classifiers }

The number of base classifiers $K$ is  important for the performance of EPD. Empirically, the best results are reached  when a minimum number for $K$ is satisfied. When $K$ surpasses this threshold no further improvement is observed.   We include examples for the $AE$ and $S$ datasets in  Figures (\ref{fig:6a}), (\ref{fig:6c}) for the Pair-0.45 configuration, and the default parameterization. In both datasets, the performance stabilizes after a certain value of $K$ (50 models for $AE$ and 100 models for $S$).

\subsubsection{Size of Evaluation Dataset}

For the creation of the $EP$, we need to separate the training dataset multiple times into a subset used for training  the base models and a subset used for evaluation, for which predictions are being made. A critical question regards how large each of these subsets will be. Using too small training subset will result in low generalization capability. On the other hand, using too large training subset, for a fixed $K$, would result in low precision in the creation of the $EP$. This could negatively impact training of $h_{Ev}$. In Figure (\ref{fig:6c})  we include the generalization performance of $h_{Ev}$ trained with EPD on the NAF:$AE$ dataset  across different evaluation- dataset sizes. We show on the x-axis the percentage of the training dataset used for evaluation. In each of the cases, the rest of the training dataset is used for training the base models. The training dataset is reshuffled per-base model. The experiment is performed with the Pair-0.45 configuration.

We notice that the performance increases substatially for smaller sizes of evaluation datasets, i.e., more data for the actual training. However, after a certain point, the performance starts to drop. We see this in Figure (\ref{fig:6c}) for evaluation size of 4\% We attribute this phenomenon to the poor formation of the $EP$ due to the small percentage of evaluation dataset in conjunction with the small $K$ for this size. We can circumvent this by increasing $K$ at the expense of additional training time.

\section{Appendix C: Architectures}
In this Appendix (Table \ref{table:Archs}) we include the architectures of the networks used in our experiments.

\section{Appendix D: Additional Results}

\begin{figure}[h]
\begin{subfigure}{0.49\columnwidth}
\includegraphics[width=\columnwidth,,height=2.5cm]{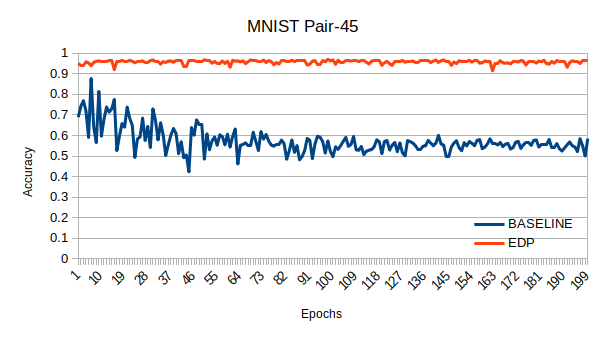}
\caption{} \label{fig:7a}
\end{subfigure}
\hspace*{\fill}
\begin{subfigure}{0.49\columnwidth}
\includegraphics[width=\columnwidth,height=2.5cm]{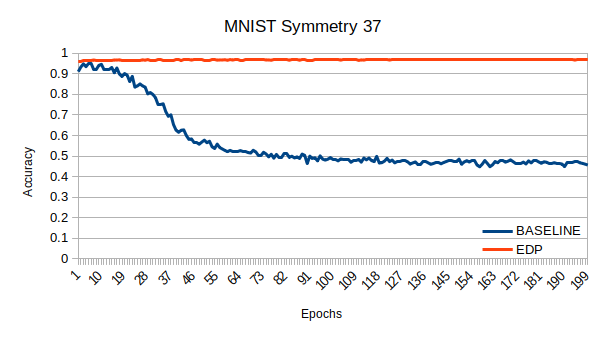}
\caption{} \label{fig:7b}
\end{subfigure}
\hspace*{\fill}

\begin{subfigure}{0.49\columnwidth}
\includegraphics[width=\columnwidth,,height=2.5cm]{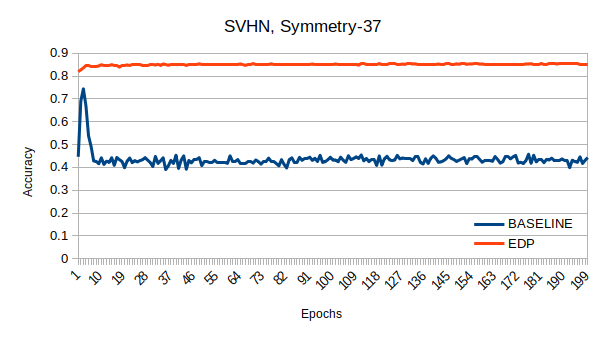}
\caption{} \label{fig:7c}
\end{subfigure}
\begin{subfigure}{0.49\columnwidth}
\includegraphics[width=\columnwidth,,height=2.5cm]{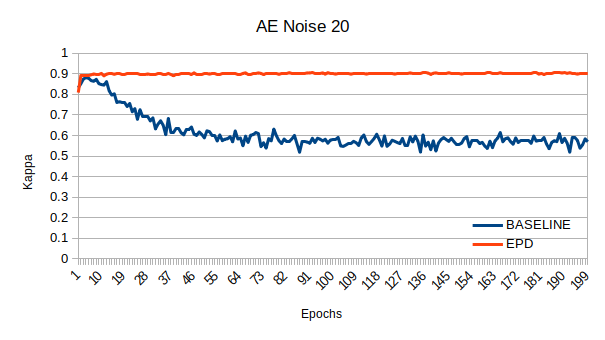}
\caption{} \label{fig:7d}
\end{subfigure}
\caption{ Various performance throughout training graphs for different noise and dataset configurations. }
\end{figure}

In this Appendix, we include additional results from our experiments. Specifically, we include the performance across all epochs for the $M$ Pair-45 (Fig. (\ref{fig:7a})), $M$ Symmetry-37 (Fig. (\ref{fig:7b})), $S$ Symmetry-37 (Fig. (\ref{fig:7c}))and $NAF:AE$ Pair-20 (Fig. (\ref{fig:7d})) configurations.

\section{Appendix E: Additional Implementation Details}

All experiments were performed on  nvidia geforce rtx 2080 ti graphics card. Python 3.6.9 and Tensorflow 1.14.0 were used.

\end{document}